%% file: main.tex
\documentclass[acmtog]{acmart}
\acmSubmissionID{754}

\usepackage{booktabs} 
\usepackage{multirow}
\usepackage{enumitem}
\usepackage{subcaption}

\usepackage[ruled]{algorithm2e} 

\SetAlFnt{\small}
\SetAlCapFnt{\small}
\SetAlCapNameFnt{\small}
\SetAlCapHSkip{0pt}

\def\shownotes{1}

\ifnum\shownotes=1
\newcommand\zhiming[1]{\textcolor{cyan}{Zhiming: #1}}
\newcommand\hl[1]{#1}

\newcommand\syn[1]{\textcolor{green}{Syn: #1}}
\newcommand\todo[1]{\textcolor{red}{#1}}
\newcommand\andreas[1]{\textcolor{orange}{Andreas: #1}}
\else 
\newcommand\andreas[1]{}
\newcommand\syn[1]{}
\newcommand\zhiming[1]{}
\newcommand\hl[1]{#1}
\newcommand\todo[1]{}
\fi

\newcommand{\methodName}{HOIGaze\xspace}

\begin{teaserfigure}
    \centering
  \includegraphics[width=0.9\textwidth]{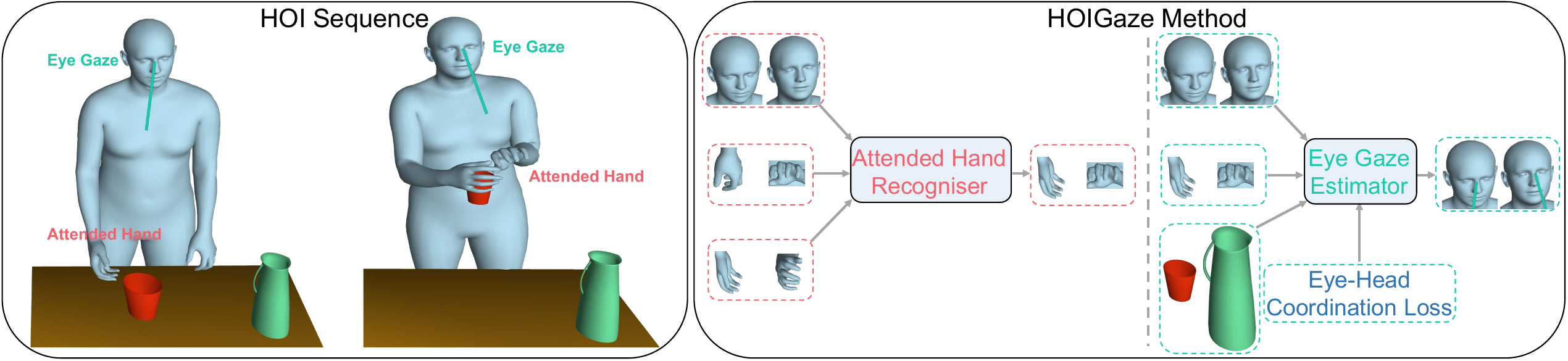}
  \caption{\methodName is a novel method for estimating eye gaze during hand-object interactions in extended reality. 
  The left figure shows an example sequence of daily HOI activity.
  \methodName uses a novel hierarchical framework that first recognises attended hand from head orientations, left and right hand gestures and then uses a gaze estimator that is trained with an eye-head coordination loss to estimate eye gaze from head orientations, attended hand, and scene objects.}
  \label{fig:teaser}
\end{teaserfigure}

\begin{document}
\title{HOIGaze: Gaze Estimation During Hand-Object Interactions in Extended Reality Exploiting Eye-Hand-Head Coordination}

\author{Zhiming Hu}
\authornote{Corresponding author}
\orcid{0000-0002-5105-9753}
\affiliation{%
  \institution{University of Stuttgart}\country{Germany}
  }
  
\author{Daniel Haeufle}
\orcid{0000-0002-3480-6892}
\affiliation{%
  \institution{University of Tuebingen}\country{Germany}  \institution{The Center for Bionic Intelligence Tuebingen Stuttgart}\country{Germany}
  }
\author{Syn Schmitt}
\orcid{0000-0002-7768-8961}
\affiliation{%
  \institution{University of Stuttgart}\country{Germany}  
  \institution{The Center for Bionic Intelligence Tuebingen Stuttgart}\country{Germany}
  }  
\author{Andreas Bulling}
\orcid{0000-0001-6317-7303}
\affiliation{%
  \institution{University of Stuttgart}\country{Germany}
  }

\input{sections/abstract}

%
%


%
%


\maketitle

\input{sections/introduction}
\input{sections/related_work}
\input{sections/method}
\input{sections/experiments}
\input{sections/application}
\input{sections/discussion}
\input{sections/conclusion}

\section*{Acknowledgement}
This work was funded, in part, by the Baden-W\"{u}rttemberg Stiftung in the scope of the AUTONOMOUS ROBOTICS project \textit{iAssistADL} granted to Syn Schmitt and Daniel Häufle.

\bibliographystyle{ACM-Reference-Format}
\bibliography{references}

\end{document}

%% file: sections/abstract.tex

\begin{abstract}
We present \methodName~-- a novel learning-based approach
for gaze estimation during hand-object interactions (HOI) in extended reality (XR).
\methodName addresses the challenging HOI setting by building on one key insight: The eye, hand, and head movements are closely coordinated during HOIs and this coordination can be exploited to identify samples that are most useful for gaze estimator training -- as such, effectively denoising the training data. 
This denoising approach is in stark contrast to previous gaze estimation methods that treated all training samples as equal.
Specifically, we propose:
1) a novel \textit{hierarchical framework} that first recognises the hand currently visually attended to and then estimates gaze direction based on the attended hand;
2) a new \textit{gaze estimator} that uses cross-modal Transformers to fuse head and hand-object features extracted using a convolutional neural network and a spatio-temporal graph convolutional network; and
3) a novel \textit{eye-head coordination loss} that upgrades training samples belonging to the coordinated eye-head movements.
We evaluate \methodName on the HOT3D and Aria digital twin (ADT) datasets and show that it significantly outperforms state-of-the-art methods, achieving an average improvement of 15.6\% on HOT3D and 6.0\% on ADT in mean angular error.
To demonstrate the potential of our method, we further report significant performance improvements for the sample downstream task of eye-based activity recognition on ADT.
Taken together, our results underline the significant information content available in eye-hand-head coordination and, as such, open up an exciting new direction for learning-based gaze estimation.
\end{abstract}

%% file: sections/introduction.tex
\section{Introduction}
With the growing popularity of extended reality (XR), analysing and understanding human behaviour in XR environments has become a popular research topic.
Human eye gaze estimation in particular has significant relevance for a number of XR applications including 1) gaze-based interaction that employs eye movements to select or interact with 3D objects~\cite{sidenmark2019eyehead}; 2) gaze-contingent rendering that maintains high rendering quality in gaze central region while reducing the fidelity in peripheral region to improve rendering efficiency~\cite{patney2016towards}; 3) gaze-based intention estimation that uses eye gaze features to predict interaction intentions~\cite{belardinelli2022intention, sun2018towards}; or 4) eye-based activity recognition that recognises user activities based on their eye movements~\cite{hu2022ehtask, jiao2024diffeyesyn}.

Estimating human eye gaze in XR environments is challenging because human gaze behaviour is influenced by both bottom-up scene content and various top-down factors, e.g. tasks that a user desires to finish~\cite{hu2021fixationnet, hu2022ehtask, jiao23supreyes}.
Prior works on gaze estimation and analysis in XR typically focused on free-viewing conditions in which no specific task is assigned to users~\cite{sitzmann2018saliency,hu2019sgaze} or non-interactive scenarios where users cannot naturally interact with the environment~\cite{hu2022ehtask,hadnett2019effect}.
However, free-viewing and non-interactive tasks have limited relevance for practical XR applications, in which users typically desire to naturally interact with the environment to perform a particular task.
Gaze estimation in the more practically relevant but also significantly more challenging scenarios that involve hand-object interactions (HOIs) has been largely neglected so far.

To fill this gap, we present \textit{\methodName}~-- the first gaze estimation method specifically geared to hand-object interactions in XR.
Our key insight is that eye, hand, and head movements are strongly coordinated during HOIs and this coordination can be used to effectively denoise the training samples to improve gaze estimation performance.
Specifically, we propose a novel \textit{hierarchical framework} that first recognises the \textit{attended hand} -- the hand that is closest to eye gaze -- by comparing the angular distance between gaze direction and a vector pointing from the eye to both hand centres.
In a second step the method then estimates eye gaze based on the attended hand (see \autoref{fig:teaser}).
We further present a new \textit{gaze estimator} that combines a convolutional neural network (CNN) to extract head features with a spatio-temporal graph convolutional network (GCN) to extract features from the attended hand gestures and scene object positions. The estimation further uses two cross-modal Transformers to fuse the head and hand-object features to estimate eye gaze.
Finally, we introduce a novel \textit{eye-head coordination loss} that upgrades training samples belonging to coordinated eye-head movements to improve the generalisation ability of the gaze estimator.
We extensively evaluate our method on the HOT3D dataset~\cite{banerjee2024introducing} for HOIs as well as on the Aria digital twin (ADT) dataset~\cite{pan2023aria} that contains a mixture of free-viewing, non-interactive, and HOI scenarios.
Experimental results show that \methodName outperforms the state of the art by a large margin, achieving an average improvement of 15.6\% on HOT3D and 6.0\% on ADT in mean angular error.
Complementing these evaluations, we also evaluate the effectiveness of our method for the sample downstream task of eye-based activity recognition on ADT and demonstrate that using our method results in significant performance improvements.
The full source code and trained models are available at zhiminghu.net/hu25\_hoigaze.

The specific contributions of our work are three-fold:
\begin{itemize}
\item We propose \methodName~-- a novel method for estimating eye gaze during HOIs in extended reality that exploits the close coordination between eye, hand, and head movements. It combines a novel hierarchical framework, a new gaze estimator that uses cross-modal Transformers to fuse the head and hand-object features extracted using a CNN and a spatio-temporal GCN, and a novel eye-head coordination loss.

\item We report extensive experiments on two public datasets for both HOI and mixed settings and demonstrate significant performance improvements over several state-of-the-art methods.

\item We show the effectiveness of our method for the sample downstream task of eye-based activity recognition, also showing significant performance improvements.
\end{itemize}

%% file: sections/related_work.tex
\section{Related Work}

\subsection{Eye Gaze Estimation}

Human eye gaze estimation or visual attention prediction has been a popular topic in the area of vision research for decades.
Typical gaze estimation methods can be classified into bottom-up methods that focus on low-level visual scene content~\cite{itti1998model, wang23_tvcg} or top-down approaches that take high-level context into consideration~\cite{wang24salchartqa, koulieris2016gaze}.
For example, Itti et al. extracted multiscale colour, intensity, and orientation features from 2D images to predict saliency map (density map of gaze distribution)~\cite{itti1998model}.
Wang et al. predicted saliency map of information visualisations using both the visualisation content and the questions assigned to the viewers~\cite{wang24salchartqa}.

Recently, with the increasing use of extended reality, a lot of efforts have been devoted to analysing and estimating human eye gaze in XR environments.
Some researchers focused on free-viewing settings where no specific task is assigned to the viewers~\cite{sitzmann2018saliency, jiao2024diffgaze}.
For example, Sitzmann et al. collected users' free-viewing eye gaze data on 360-degree images and adapted existing saliency predictors to predict saliency maps of 360-degree images~\cite{sitzmann2018saliency}.
Hu et al. recorded users' eye movements during freely exploring static or dynamic virtual environments for developing eye gaze estimation models~\cite{hu2019sgaze, hu2020dgaze}.
Other researchers devoted to analysing eye gaze in non-interactive scenarios where users cannot naturally interact with the environment.
Specifically, Hadnett-Hunter et al. explored the effect of three tasks on visual attention in desktop monitor-based virtual environments~\cite{hadnett2019effect}.
Hu et al. analysed the differences of eye gaze patterns under four different visual tasks during viewing 360-degree videos~\cite{hu2022ehtask}.
However, free-viewing or non-interactive settings have limited relevance for practical XR applications.
In stark contrast, in this work we investigate gaze estimation in the more challenging but also more practically relevant hand-object interaction scenarios.

\subsection{Eye-Hand-Head Coordination}

Analysing and understanding the coordination of human eye, hand, and head movements is a significant topic in the areas of cognitive science and human-centred computing.
Stahl~\cite{stahl1999amplitude} and Sidenmark et al.~\cite{sidenmark2019eye} both found that eye gaze is coordinated with head movements during the gaze shift process.
Hu et al. revealed that human eye movements in virtual environments have strong correlations with their head movements in both free-viewing~\cite{hu2019sgaze,hu2020dgaze} and task-oriented scenarios~\cite{hu2021fixationnet,hu2022ehtask}.
Kothari et al. observed coordinated patterns of human eye and head movements in real-world daily activities~\cite{kothari2020gaze}.
Hu et al. learned generalisable joint representations of hand trajectories and head orientations in extended reality~\cite{hu2024haheae}.
Belardinelli et al. observed the coordinated patterns of human eye gaze and hand trajectories during daily pick and place activities in virtual environments
~\cite{belardinelli2022intention}.
Hu et al. revealed the correlation between eye gaze direction and wrist movements in various daily activities~\cite{hu24pose2gaze}.
In stark contrast with prior works, we are the first to exploit eye-hand-head coordination to effectively denoise training samples to improve gaze estimation performance.

\subsection{Hand-Object Interaction}

Hand-object interaction is an important interaction paradigm in people's daily life and has been studied by many researchers.
Damen et al. employed egocentric images in various HOI scenarios to recognise or anticipate user activities~\cite{damen2022rescaling} while Zhang et al. used egocentric images with per-pixel segmentation labels of hands and objects for activity recognition~\cite{zhang2022fine}.
Shi et al. took temporal inter-dependencies between HOI actions into consideration to generate procedure actions in instructional videos~\cite{shi25_wacv}.
Liu et al. explored action recognition, motion forecasting, and cooperative grasp synthesis during bimanual hand-object manipulation process~\cite{liu2024taco}.
Zhan et al. investigated hand mesh reconstruction, task-aware motion fulfillment, and complex task completion during complex HOI activities~\cite{zhan2024oakink2}.
Despite plenty of research on HOIs, eye gaze estimation during HOI activities has been neglected so far.
To fill this gap, in this work we explore gaze estimation under HOI scenarios.

%% file: sections/method.tex
\section{Method}

\subsection{Method Design}

\begin{figure*}[t]
    \centering
    \includegraphics[width=0.9\textwidth]{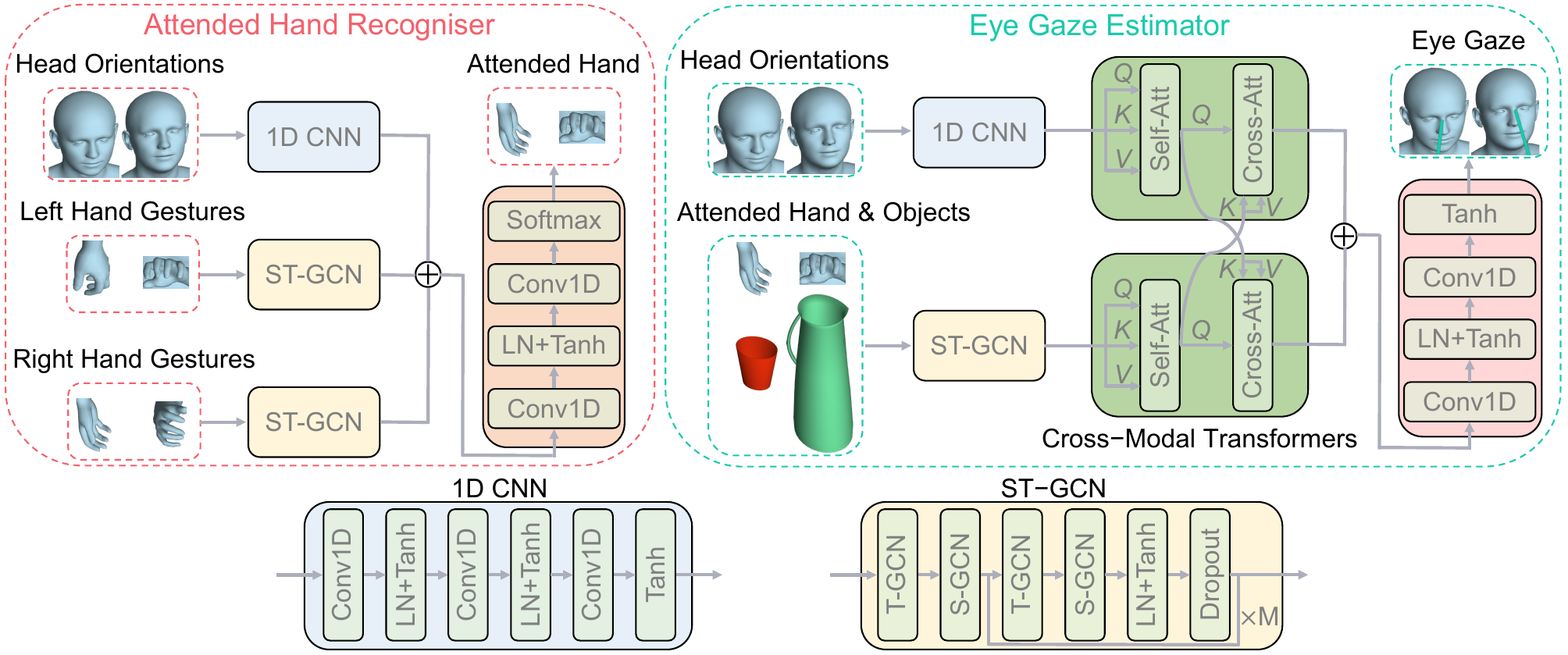}
    \caption{\methodName combines an attended hand recogniser and an eye gaze estimator. The attended hand recogniser uses a 1D CNN and two ST-GCNs to extract features from head orientations, left and right hand gestures, respectively, to recognise the attended hand. The gaze estimator uses an ST-GCN to extract features from the attended hand and scene objects, and then fuses the head and hand-object features using cross-modal Transformers to estimate eye gaze.}\label{fig:method}
\end{figure*}

\paragraph{Problem Formulation} 
We define gaze estimation during hand-object interactions in extended reality as the task of generating a sequence of eye gaze directions $G = \{g_i\}_{i=1}^T\in R^{3\times T}$, where $g_i$ is a 3D unit vector and $T$ is the sequence length, from hand gestures, scene objects, and head movements.
To enhance hand gestures with more context, we represent hand gestures using the 3D positions of all the hand joints as well as the head and wrist positions.
Specifically, the left and right hand gestures are represented as $LH = \{he_i, lw_i, rw_i, lh_i\}_{i=1}^T\in R^{3\times (N+3)\times T}$ and $RH = \{he_i, lw_i, rw_i, rh_i\}_{i=1}^T\in R^{3\times (N+3)\times T}$ respectively, where $he_i$, $lw_i$, and $rw_i$ are the 3D positions of the head, left and right wrists, $lh_i\in R^{3\times N}$ and $rh_i\in R^{3\times N}$ refer to the 3D positions of the left and right hands and $N$ is the number of hand joints.
We denote scene objects using the 3D positions of the object centres $O = \{o^1_i, o^2_i, ..., o^J_i\}_{i=1}^T\in R^{3\times J\times T}$, where $J$ is the number of objects.
We use the head forward directions to represent head orientations $H = \{h_i\}_{i=1}^T\in R^{3\times T}$, where $h_i$ is a 3D unit vector.

\paragraph{Design of \methodName}
We observe that during hand-object interactions human visual attention is usually attracted to one hand at a specific time.
For example, in a scenario where a user first picks up a cup on the table using their right hand and then picks up a jug with their left hand, the user would pay their attention to the right hand first and then turn to the left hand (see \autoref{fig:teaser}).
The attended hand is highly correlated with eye gaze while the unattended hand has little correlation.
Therefore, knowing which hand is the attended one has significant potential for estimating eye gaze.
Based on this observation, we propose a novel hierarchical framework that combines an attended hand recogniser and an eye gaze estimator (see \autoref{fig:method} for an overview of our method).
The attended hand recogniser uses a convolutional neural network and two spatio-temporal graph convolutional networks to extract features from head orientations, left and right hand gestures respectively, and then concatenates these features to recognise attended hand via a convolutional neural network.
The gaze estimator first uses a convolutional neural network to extract head features and a spatio-temporal graph convolutional network to extract features from the attended hand gestures and scene object positions, then employs two cross-modal Transformers to fuse the head and hand-object features, and finally uses a convolutional neural network to estimate eye gaze from the fused features.

\subsection{Attended Hand Recogniser}
\label{sec:attended_hand_recogniser}

\paragraph{Head Orientation Feature Extraction}
Considering the good performance of 1D CNN for processing head movement data~\cite{hu2020dgaze,hu2021fixationnet, hu2022ehtask}, we used three 1D CNN layers, each with $32$ channels and a kernel size of three, to extract features from head orientations $H\in R^{3\times T}$.
The first two CNN layers were followed by a layer normalisation (LN) and a Tanh activation while the third CNN layer was followed by a Tanh activation.
After the three CNN layers, we obtained the head orientation features $f_{he}\in R^{32\times T}$.

\paragraph{Hand Gesture Feature Extraction}
In light of the superior performance of graph convolutional networks for processing body and hand pose data~\cite{hu24hoimotion, tang2024prompting}, we used two spatio-temporal GCNs to extract features from the left and right hand gestures respectively.
Specifically, we modelled the hand gesture data ($LH\in R^{3\times (N+3)\times T}$ or $RH\in R^{3\times (N+3)\times T}$) as fully connected spatial and temporal graphs with their adjacency matrices measuring the weights between each pair of nodes.
The spatial graph consists of $N+3$ joints representing the hand joints, head, and wrists, respectively, while the temporal graph contains $T$ nodes corresponding to hand gesture at different time steps.
We first mapped the original hand gesture data into a latent feature space using a spatial-temporal graph convolutional network (ST-GCN).
The ST-GCN first multiplied the data with a temporal adjacency matrix $A^T \in R^{T\times T}$ to perform temporal convolution, then used a feature matrix $W\in R^{3\times8}$ to map the original node features ($3$ dimensions) into latent space ($8$ dimensions), and finally multiplied the data with a spatial adjacency matrix $A^S \in R^{(N+3)\times (N+3)}$ to perform spatial convolution.
We copied the output of the ST-GCN along the temporal dimension ($R^{8\times(N+3)\times T}\rightarrow R^{8\times(N+3)\times 2T}$) to enhance the features~\cite{ma2022progressively}.
We further used a residual GCN module that contains two GCN blocks to process the enhanced data.
Each GCN block consists of an ST-GCN, a layer normalisation, a Tanh activation, and a dropout layer with a dropout rate of $0.3$ to avoid overfitting.
The feature matrix of the ST-GCN used in the GCN block was set to $W\in R^{8\times8}$, ensuring that the input and output of the GCN block had the same size.
A residual connection was applied for each GCN block to improve the network flow.
We finally cut the output of the residual GCN module in half along the temporal dimension to obtain the hand gesture features ($f_{lh}\in R^{8\times(N+3)\times T}$ and $f_{rh}\in R^{8\times(N+3)\times T}$).

\paragraph{Attended Hand Recognition}
To recognise attended hand, we first aggregated the hand gesture features along the spatial dimension ($R^{8\times (N+3)\times T}\rightarrow R^{8(N+3)\times T}$).
We then concatenated the head orientation, left and right hand gesture features along the spatial dimension and obtained $f\in R^{(16(N+3)+32)\times T}$.
We finally applied two CNN layers, each with a kernel size of three, to process the concatenated features.
The first CNN layer had $64$ channels and was followed by a layer normalisation and a Tanh activation function while the second CNN layer had two channels and was followed by a Softmax activation to generate the probabilities of the left and right hands being the attended hand.

\subsection{Eye Gaze Estimator}\label{sec:gaze_estimator}

\paragraph{Head Orientation Feature Extraction}
We employed the same CNN module as used in the attended hand recogniser (\autoref{sec:attended_hand_recogniser}) to obtain head orientation features $f_{he}\in R^{32\times T}$.

\paragraph{Hand-Object Feature Extraction}
Considering that human visual attention is more likely to be attracted by the scene objects that are close to the attended hand, we first calculated the average distance between object centre and all the joints of the attended hand for every scene object and then added the nearest scene object to the representation of the attended hand $AH = \{he_i, lw_i, rw_i, ah_i, o^{ah}_i\}_{i=1}^T\in R^{3\times (N+4)\times T}$, where $ah_i$ refers to the attended hand joints and $o^{ah}_i$ denotes the 3D positions of the nearest scene object.
We further used an ST-GCN and a residual GCN module that contains four GCN blocks to extract features from the hand-object data.
The ST-GCN and GCN block had the same architecture as used in the attended hand recogniser (\autoref{sec:attended_hand_recogniser}) except that the spatial graph had $N+4$ nodes rather than $N+3$.
We finally aggregated the hand-object features along the spatial dimension and obtained $f_{ah}\in R^{8(N+4)\times T}$.

\paragraph{Head-Hand-Object Feature Fusion}
In light of the good performance of cross-modal transformers for fusing different features~\cite{zhang2022motiondiffuse, yan24gazemodiff}, we used cross-modal transformers to fuse the head and hand-object features by modelling correlations between different time steps.
To this end, we first applied a self-attention block to enhance the head and hand-object features respectively.
Given input features $X\in R^{T\times n}$ where $T$ refers to sequence length and $n$ denotes the feature dimension, the self-attention block first calculated query feature vectors $Q\in R^{T\times n}$, key feature vectors $K\in R^{T\times n}$, and value feature vectors $V\in R^{T\times n}$ using $Q=W_qX$, $K=W_kX$, and $V=W_vX$, where $W_q$, $W_k$, and $W_v$ are the linear projections to generate $Q$, $K$, and $V$.
The self-attention block then enhanced the input features $X$ using
\begin{equation} \label{eq:attention}
  Y = X + softmax(\frac{Q\otimes K^T}{\sqrt{n}})\otimes V,
\end{equation} 
where $Y\in R^{T\times n}$ is the enhanced features and $\otimes$ refers to matrix multiplication.
After the self-attention block, we further used a cross-attention block to fuse the head and hand-object features.
Specifically, we used one modality to calculate $Q$ and the other modality to compute $K$ and $V$ and then applied \autoref{eq:attention} to fuse the two modalities.
After the cross-modal transformers, we obtained the enhanced head features $f^{'}_{he}\in R^{32\times T}$ and hand-object features $f^{'}_{ah}\in R^{8(N+4)\times T}$.

\paragraph{Eye Gaze Estimation}
To estimate eye gaze, we first concatenated the head and hand-object features along the spatial dimension and obtained $f\in R^{(8(N+4)+32)\times T}$.
We then used two 1D CNN layers, each with a kernel size of three, to process the concatenated features.
The first CNN layer had $64$ channels and was followed by a layer normalisation and a Tanh activation while the second CNN layer used three channels and a Tanh activation to generate eye gaze.
We finally normalised the output to unit vectors to represent eye gaze directions $\hat{G} = \{\hat{g}_i\}_{i=1}^T\in R^{3\times T}$.

\subsection{Loss Function}
We first trained the attended hand recogniser and then used the recognised attended hand to train the gaze estimator.
Specifically, we trained the recogniser using the cross entropy loss given its good performance for classification task.
To train the gaze estimator, we proposed a novel eye-head coordination loss that increases the weights of the training samples belonging to eye-head coordinated movements:
\begin{equation}
\ell_i = \left\{
         \begin{array}{lr}
         f_{eh}*(g_i - \hat{g}_i)^2,  & \text{if } g_i\cdot h_i > Cos_{eh}\\
         (g_i - \hat{g}_i)^2,  & \text{otherwise}\\ 
             \end{array}
\right.
\end{equation}
where $g_i\cdot h_i$ calculates the cosine similarity between eye gaze direction and head orientation, $Cos_{eh}$ is the threshold for eye-head cosine similarity and is set to $0.8$, and $f_{eh}$ is the weighting factor and is set to $4.0$.
The insight behind this loss function is that the coordinated eye-head movements are much more pervasive than the movements with little eye-head correlation~\cite{hu24pose2gaze,sitzmann2018saliency,nakashima2015saliency, sidenmark2019eye}.
Therefore, increasing the weights of coordinated eye-head training samples can improve the generalisation ability of the gaze estimator.

%% file: sections/experiments.tex
\section{Experiments and Results} \label{sec:experiments}


\subsection{Datasets} \label{sec:experimental_datasets}

To evaluate our method's generalisation capability for different scenarios, we tested our method on the HOT3D dataset~\cite{banerjee2024introducing} for HOI setting as well as on the ADT dataset~\cite{pan2023aria} for mixed setting that contains a mixture of free-viewing, non-interactive, and HOI scenarios.

\paragraph{HOT3D Dataset}
The HOT3D dataset contains human eye gaze, head pose, wrist pose, \hl{gestures of $20$ hand joints}, and 3D scene objects recorded at $30$ Hz during various HOI activities in three different environments including \textit{living room}, \textit{kitchen}, and \textit{office}.
The original test set of HOT3D is not publicly available, so we only used HOT3D's original training set that contains $136$ recordings from nine subjects and each recording lasts for around two minutes.
To evaluate our method's generalisation capability for different users and environments, we respectively performed a cross-user evaluation and a cross-scene evaluation.
For cross-user evaluation, we split the data into three user sets, i.e. \{P1, P2, P3\}, \{P9, P10, P11\}, and \{P12, P14, P15\}, trained on two sets and tested on the remaining one, and repeated this procedure three times by testing for a different user set.
For cross-scene evaluation, we trained on two environments and tested on the remaining one, and repeated this procedure three times by testing for a different scene.

\paragraph{ADT Dataset}
The ADT dataset collects human eye gaze, head pose, wrist pose and 3D scene objects at $30$ Hz during various indoor activities including \textit{work}, \textit{room decoration}, and \textit{meal preparation}, in which free-viewing, non-interactive, and HOI scenarios are mixed together.
The dataset contains $34$ sequences and each sequence lasts for around two minutes.
For evaluation on ADT, we followed prior works~\cite{hu24pose2gaze, hu24hoimotion} to use $24$ sequences for training and the remaining $10$ sequences for testing.
The ADT dataset does not record dynamic hand gestures but provides a static \hl{gesture of $15$ hand joints}, in which the 3D positions of the fingers are determined only by the wrist pose while the relative finger pose doesn't change over time.
For experiments on ADT, we used the static hand gesture to train and test our method.

\subsection{Evaluation Settings}

\paragraph{Evaluation Metric}
As is common in gaze estimation~\cite{hu24pose2gaze, hu2021fixationnet, hu2020dgaze}, we used the mean angular error between the estimated gaze directions and the ground truth as the metric to evaluate different methods.

\paragraph{Baselines}
We compared our method with the following state-of-the-art gaze estimation methods designed for XR environments:

\begin{itemize}[noitemsep,leftmargin=*]    
    \item \textit{Head Direction}: \textit{Head Direction} is frequently used as a proxy for eye gaze in XR due to the strong correlation between eye and head movements~\cite{sitzmann2018saliency,hu2019sgaze,hu2020dgaze}.
    
    \item \textit{DGaze}~\cite{hu2020dgaze}: \textit{DGaze} estimates eye gaze from the scene content and head movements via convolutional neural networks.    
    
    \item \textit{FixationNet}~\cite{hu2021fixationnet}: \textit{FixationNet} combines prior knowledge of gaze distribution with head and scene features extracted by convolutional neural networks to estimate eye gaze.    
    
    \item \textit{Pose2gaze}~\cite{hu24pose2gaze}: \textit{Pose2gaze} estimates eye gaze from body movements using graph convolutional networks.
\end{itemize}

\paragraph{Time Horizon}
We used $15$ frames (corresponding to $500$ ms) of hand-head-object data as input to estimate the corresponding gaze directions $G_{t:t+14} = \{g_{t}, g_{t+1}, ..., g_{t+14}\}$ following common evaluation settings for gaze estimation in XR~~\cite{hu24pose2gaze, hu2020dgaze}.

\paragraph{Implementation Details}
We trained the baseline methods from scratch using their default parameters.
We trained our attended hand recogniser using the AdamW optimiser with an initial learning rate of $0.005$ and a weight decay coefficient of $0.05$.
We decayed the learning rate by $0.95$ every epoch and trained the recogniser for $60$ epochs using a batch size of $32$.
We trained our gaze estimator using the Adam optimiser with an initial learning rate of $0.005$ that was decayed by $0.95$ every epoch.
We used a batch size of $32$ to train the gaze estimator for a total of $80$ epochs.
We implemented our method with the PyTorch framework using a Linux machine with one NVIDIA V100 GPU.

\subsection{Gaze Estimation Results} \label{sec:gaze_estimation_results}

\begin{table*}[t]
	\centering
	\caption{Mean angular errors of different methods on the HOT3D and ADT datasets. Best results are in bold.}\label{tab:results}
        \resizebox{\textwidth}{!}{
	\begin{tabular}{ccccccccccccc}
		\toprule
    &\multicolumn{4}{c}{\textbf{HOT3D (Cross-User)}} &\multicolumn{4}{c}{\textbf{HOT3D (Cross-Scene)}} &\multicolumn{4}{c}{\textbf{ADT}}\\ \cmidrule(lr){2-5} \cmidrule(lr){6-9} \cmidrule(lr){10-13}
    &\textit{\{P1, P2, P3\}} &\textit{\{P9, P10, P11\}} &\textit{\{P12, P14, P15\}} &Average &\textit{Room} &\textit{Kitchen} &\textit{Office} &Average &\textit{Work} &\textit{Decoration} &\textit{Meal} &Average\\ \hline
  \textit{Head Direction} & $23.24^\circ$ & $28.00^\circ$ & $17.85^\circ$ & $23.20^\circ$ & $23.69^\circ$ & $22.83^\circ$ & $23.16^\circ$ & $23.20^\circ$ & $22.88^\circ$ & $18.44^\circ$ & $25.23^\circ$ & $22.25^\circ$				
  \\ 
  \textit{DGaze}~\cite{hu2020dgaze}& $12.17^\circ$ & $15.08^\circ$ & $14.87^\circ$ & $14.29^\circ$ & $13.37^\circ$ & $12.98^\circ$ & $11.39^\circ$ & $12.81^\circ$ & $8.84^\circ$ & $10.53^\circ$ & $10.77^\circ$ & $9.92^\circ$\\ 
  \textit{FixationNet}~\cite{hu2021fixationnet} & $11.90^\circ$ & $14.60^\circ$ & $14.78^\circ$ & $14.00^\circ$ & $12.78^\circ$ & $12.84^\circ$ & $11.34^\circ$ & $12.53^\circ$ & $8.82^\circ$ & $10.50^\circ$ & $10.83^\circ$ & $9.92^\circ$		  
  \\  
  \textit{Pose2Gaze}~\cite{hu24pose2gaze} & $10.69^\circ$ & $10.73^\circ$ & $11.80^\circ$ & $11.10^\circ$ & $9.79^\circ$ & $9.73^\circ$ & $9.96^\circ$ & $9.80^\circ$ & $8.25^\circ$ & $9.71^\circ$ & $10.43^\circ$ & $9.34^\circ$\\		  
  \midrule    
    Ours & $\textbf{9.23}^\circ$ & $\textbf{9.16}^\circ$ & $\textbf{9.69}^\circ$ & $\textbf{9.37}^\circ$ & $\textbf{8.55}^\circ$ & $\textbf{8.69}^\circ$ & $\textbf{8.69}^\circ$ & $\textbf{8.64}^\circ$ & $\textbf{7.81}^\circ$ & $\textbf{9.46}^\circ$ & $\textbf{9.41}^\circ$ & $\textbf{8.78}^\circ$\\  
  Ours w/o attended hand& $9.89^\circ$ & $11.24^\circ$ & $10.57^\circ$ & $10.67^\circ$ & $9.71^\circ$ & $9.32^\circ$ & $9.16^\circ$ & $9.43^\circ$ & $8.26^\circ$ & $9.97^\circ$ & $9.87^\circ$ & $9.25^\circ$\\			  
  Ours w/o Transformers& $9.60^\circ$ & $10.07^\circ$ & $10.24^\circ$ & $10.02^\circ$ & $8.87^\circ$ & $8.85^\circ$ & $9.17^\circ$ & $8.92^\circ$ & $8.03^\circ$ & $9.74^\circ$ & $9.96^\circ$ & $9.12^\circ$\\		  
  Ours w/o eye-head coord. loss& $9.83^\circ$ & $9.48^\circ$ & $9.70^\circ$ & $9.64^\circ$ & $8.79^\circ$ & $8.71^\circ$ & $8.84^\circ$ & $8.76^\circ$ & $7.87^\circ$ & $9.49^\circ$ & $9.71^\circ$ & $8.90^\circ$\\    
  \midrule    
  Ours w/ GT attended hand & $\textbf{8.68}^\circ$ & $\textbf{8.82}^\circ$ & $\textbf{9.35}^\circ$ & $\textbf{8.98}^\circ$ & $\textbf{8.41}^\circ$ & $\textbf{8.25}^\circ$ & $\textbf{8.40}^\circ$ & $\textbf{8.34}^\circ$ & $\textbf{7.59}^\circ$ & $\textbf{9.18}^\circ$ & $\textbf{9.28}^\circ$ & $\textbf{8.57}^\circ$\\ 		  
        \bottomrule
	\end{tabular}}
\end{table*}

\begin{figure*}[t]
    \centering
    \subfloat[HOT3D (Cross-User)]{\includegraphics[width=0.33\linewidth]{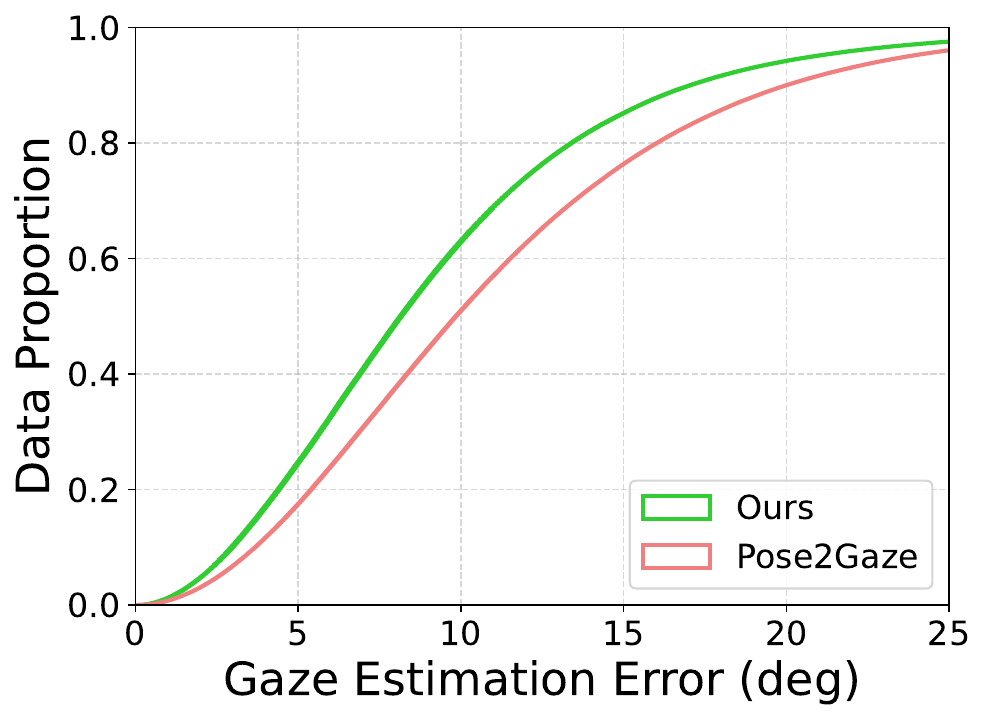}} 
    \subfloat[HOT3D (Cross-Scene)]{\includegraphics[width=0.33\linewidth]{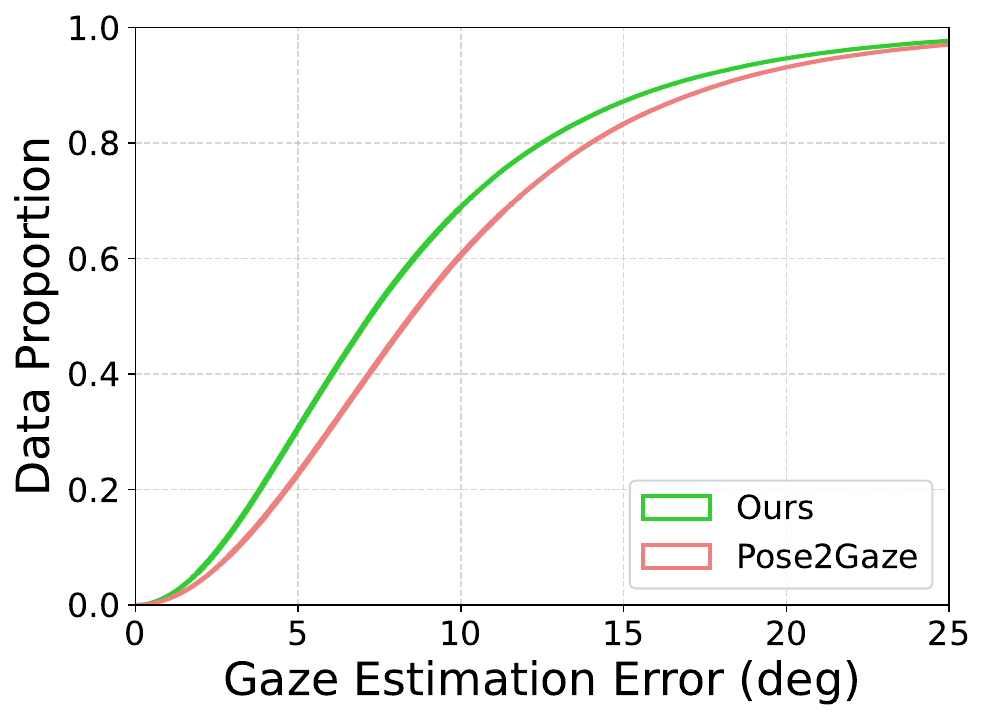}}
    \subfloat[ADT]{\includegraphics[width=0.33\linewidth]{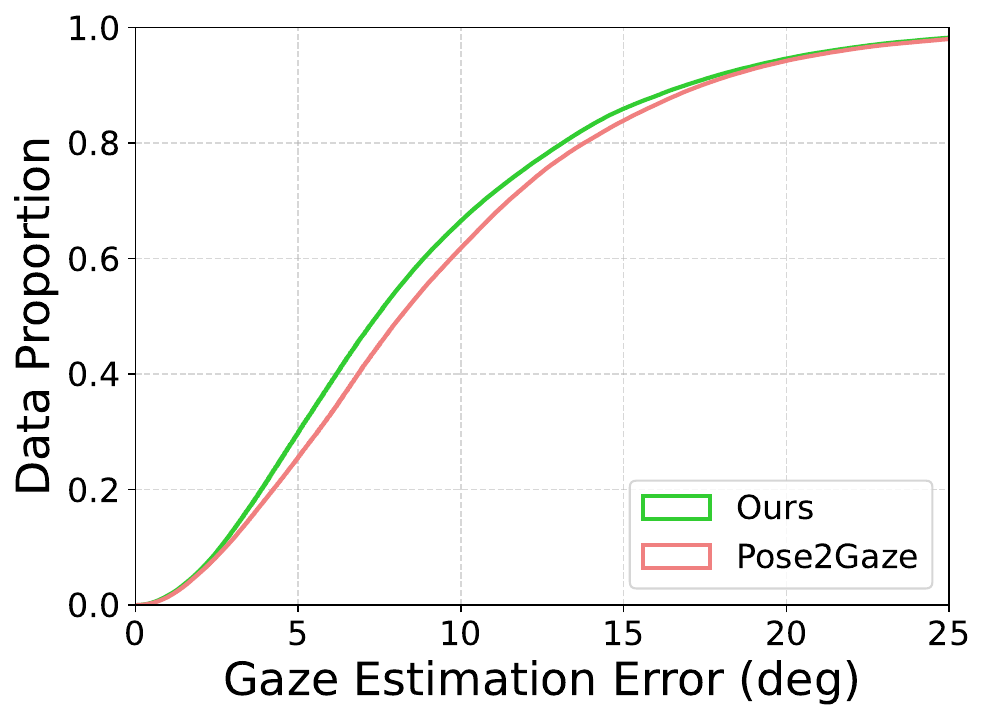}}
    \caption{The cumulative distribution functions of different methods' estimation errors on HOT3D (Cross-User), HOT3D (Cross-Scene), and ADT. The higher the CDF curve, the better the performance. Our method achieves better performance than other methods in terms of estimation error distributions.} \label{fig:error_distribution}
\end{figure*}

\begin{figure*}[t]
    \centering
    \includegraphics[width=\textwidth]{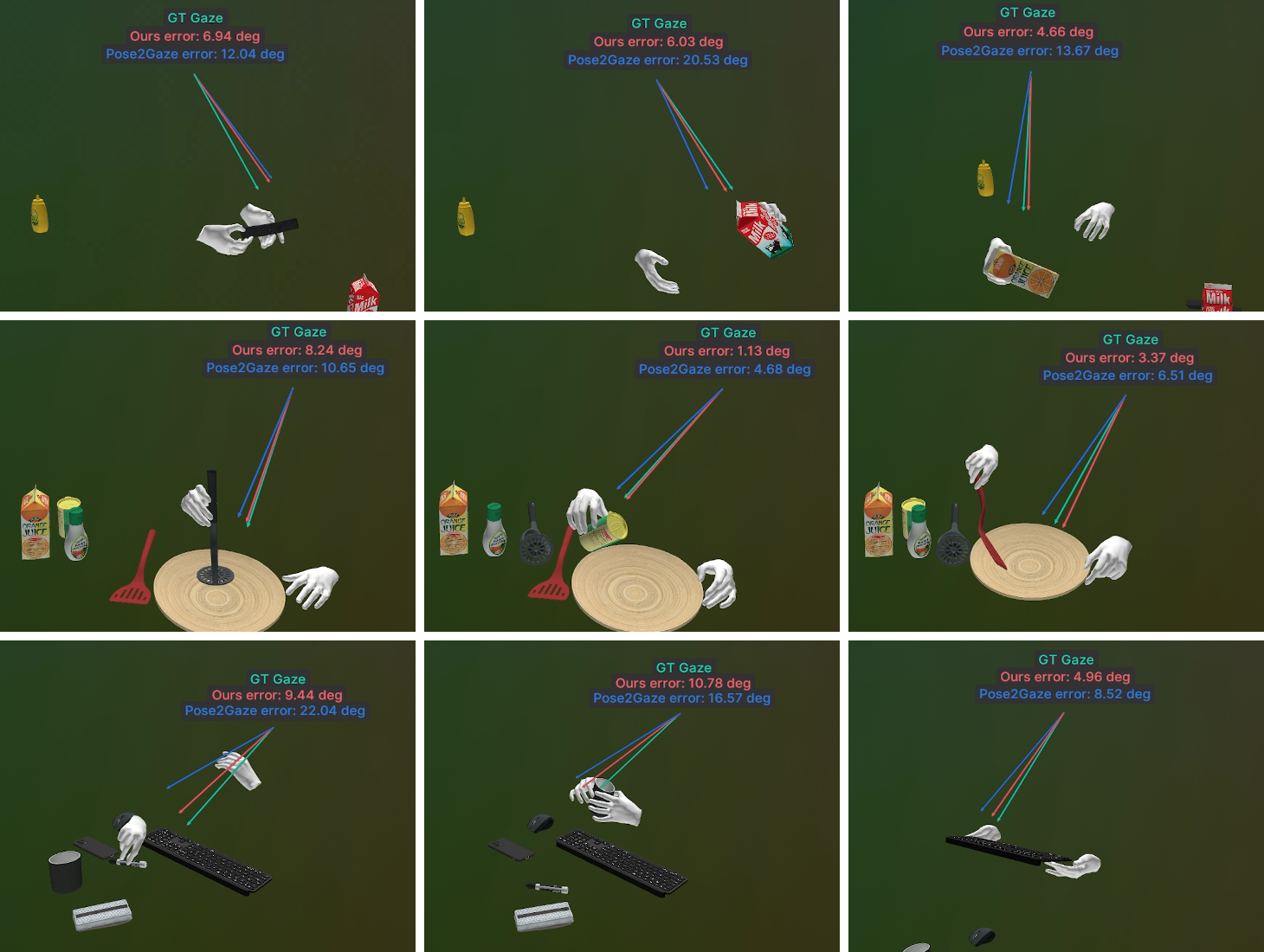}
    \caption{Visualisation of the gaze estimation results from our method and the state-of-the-art method \textit{Pose2Gaze}~\cite{hu24pose2gaze} on the HOT3D dataset. The green arrow represents the ground truth eye gaze, the red arrow denotes our method, the blue arrow refers to \textit{Pose2Gaze}. Our method exhibits higher estimation accuracy than the state-of-the-art method at different scenarios and different activities.}\label{fig:results}
\end{figure*}

\paragraph{Results on HOT3D (Cross-User)}
\autoref{tab:results}\textit{-HOT3D (Cross-User)} summarises the performances of different methods on the HOT3D dataset for cross-user evaluation.
We can see that our method consistently outperforms the state of the art in terms of both average performance and performances at different user sets.
Specifically, our method achieves an average improvement of $15.6\%$ ($9.37^\circ$ \textit{vs.} $11.10^\circ$) in mean angular error, an improvement of $13.7\%$ ($9.23^\circ$ \textit{vs.} $10.69^\circ$) on \{P1, P2, P3\}, $14.6\%$ ($9.16^\circ$ \textit{vs.} $10.73^\circ$) on \{P9, P10, P11\}, and $17.9\%$ ($9.69^\circ$ \textit{vs.} $11.80^\circ$) on \{P12, P14, P15\}.
We further performed a paired Wilcoxon signed-rank test to compare the mean angular error of our method with that of the state-of-the-art methods and the results validated that the differences between our method and prior methods are statistically significant ($p<0.01$).
\hl{We also analysed the cumulative distribution functions (CDFs) of different methods' estimation errors and validated that our method achieves better performance than other methods (see \autoref{fig:error_distribution}).}
\autoref{fig:results} shows the visualisation of the gaze estimation results from our method and the state-of-the-art method \textit{Pose2Gaze}~\cite{hu24pose2gaze}.
We can see that our method achieves better performance than the state-of-the-art method at different scenarios and different activities.
More visualisation results are provided in the supplementary video.

\paragraph{Results on HOT3D (Cross-Scene)}
\autoref{tab:results}\textit{-HOT3D (Cross-Scene)} shows the mean angular errors of different methods on HOT3D for cross-scene evaluation.
It can be seen that our method consistently outperforms other methods in both average performance and performances at different environments.
Specifically, our method achieves an average improvement of $11.8\%$ ($8.64^\circ$ \textit{vs.} $9.80^\circ$), an improvement of $12.7\%$ ($8.55^\circ$ \textit{vs.} $9.79^\circ$) at \textit{living room}, $10.7\%$ ($8.69^\circ$ \textit{vs.} $9.73^\circ$) at \textit{kitchen}, and $12.8\%$ ($8.69^\circ$ \textit{vs.} $9.96^\circ$) at \textit{office}.
A paired Wilcoxon signed-rank test was conducted and the results indicated that the differences between our method and the state of the art are statistically significant ($p<0.01$).
\hl{The results in \autoref{fig:error_distribution} further show that our method outperforms other methods in terms of estimation error distributions.}

\paragraph{Results on ADT}
The gaze estimation performances of different methods on the ADT dataset are presented at \autoref{tab:results}\textit{-ADT}.
We can see from the table that our method consistently outperforms prior methods at different activities including \textit{work}, \textit{room decoration}, and \textit{meal preparation}, and achieves an average improvement of $6.0\%$ ($8.78^\circ$ \textit{vs.} $9.34^\circ$).
A paired Wilcoxon signed-rank test validated that our improvement is statistically significant ($p<0.01$).
\hl{We further demonstrate that our method achieves better performance than other methods in estimation error distributions (see \autoref{fig:error_distribution}).}

\subsection{Ablation Study}
\paragraph{Attended Hand}
To test the effectiveness of the recognised attended hand, we re-trained our gaze estimator using both the left and right hands rather than the attended hand.
We can see from \autoref{tab:results} that our method significantly outperforms the ablated version of not using attended hand (paired Wilcoxon signed-rank test, $p<0.01$).
In addition, we also re-trained our gaze estimator using the ground truth attended hand and the results in \autoref{tab:results} further validate the usefulness of the attended hand.
Furthermore, we analysed the error cases when the recognised attended hand is wrong and found that our method can achieve superior or comparable performance with the state of the art in these cases, demonstrating the robustness of our method (see supplementary material for details).

\paragraph{Cross-Modal Transformer}
We re-trained our gaze estimator without using the cross-modal Transformers and the results in \autoref{tab:results} demonstrate that cross-modal Transformers help improve the performance significantly (paired Wilcoxon signed-rank test, $p<0.01$).
 We also validated that both the self-attention and cross-attention blocks used in the cross-modal Transformers contribute to the overall performance (see supplementary material for details).
 
\paragraph{Eye-Head Coordination Loss}
We replaced the eye-head coordination loss with a mean squared error (MSE) loss to re-train our method.
The results in \autoref{tab:results} verify that the eye-head coordination loss can significantly improve our method's gaze estimation performance (paired Wilcoxon signed-rank test, $p<0.01$).

\paragraph{GCN in Attended Hand Recogniser and Gaze Estimator}
We respectively removed the residual GCNs used in our attended hand recogniser and gaze estimator to re-train our method.
We can see from the results in \autoref{tab:ablation} that using residual GCNs achieves significantly better performances than not using them (paired Wilcoxon signed-rank test, $p<0.01$).
We also changed the number of GCN layers and validated that using two residual GCNs in the attended hand recogniser and four residual GCNs in the gaze estimator achieves the best performance (see supplementary material for details).

\paragraph{Input Modality}
We respectively tested different ablated versions of our method that did not contain head orientations, head-wrist positions, dynamic hand gestures (i.e., replace dynamic hand gestures with static ones), and scene objects.
We can see from \autoref{tab:ablation} that our method consistently outperforms the ablated versions and the results are statistically significant (paired Wilcoxon signed-rank test, $p<0.01$), thus underlining the effectiveness of each input modality used in our method. 
We also changed the number of scene objects and validated that using the nearest scene object achieves the best performance (see supplementary material for details).

\begin{table}[t] 
	\centering
	\caption{Mean angular errors of our method's different ablated versions on the HOT3D and ADT datasets. Best results are in bold.} \label{tab:ablation} 
        \resizebox{0.5\textwidth}{!}{
	\begin{tabular}{cccc}
		\toprule
		      & \textbf{HOT3D-User} & \textbf{HOT3D-Scene} & \textbf{ADT} \\ \hline
    w/o recogniser GCN & $9.75^\circ$ & $8.78^\circ$ & $8.99^\circ$\\    
    w/o estimator GCN & $9.87^\circ$ & $9.13^\circ$ & $9.20^\circ$\\    
    w/o head orientations & $10.53^\circ$ & $9.29^\circ$ & $9.42
^\circ$ \\ 
    w/o head-wrist positions & $12.82^\circ$ & $11.47^\circ$ & $9.57^\circ$ 
                \\ 
    w/o hand gestures & $9.63^\circ$ & $8.74^\circ$ & - 
                \\ 
    w/o scene objects & $10.34^\circ$ & $9.19^\circ$ & $9.03^\circ$ \\ 
        Ours & $\textbf{9.37}^\circ$ & $\textbf{8.64}^\circ$ & $\textbf{8.78}^\circ$\\
        \bottomrule
	\end{tabular}}
\end{table}

%% file: sections/application.tex
\section{Eye-based Activity Recognition} \label{sec:application}

Activity recognition is important for many XR scenarios such as low-latency predictive interfaces~\cite{david2021towards, keshava2020decoding}, adaptive virtual environment design~\cite{hadnett2019effect}, or human-aware intelligent systems \cite{vortmann2020attention}.
It is well-known that human eye gaze can be directly used to recognise user activities~\cite{hu2022ehtask, bulling2010eye, coutrot2018scanpath}.
Therefore, eye-based activity recognition is a particularly relevant sample downstream task to further evaluate the quality of the estimated eye gaze.

\paragraph{Dataset}
We tested on the ADT dataset given that it provides activity labels for the recorded sequences.
We used the same training and test sets as described in \autoref{sec:experimental_datasets}.

\paragraph{Activity Recognition Method}
We used \textit{EHTask}~\cite{hu2022ehtask} -- the state-of-the-art eye-based activity recogniser to evaluate the estimated eye gaze.
\textit{EHTask} employs a 1D CNN and a bidirectional gated recurrent unit (GRU) to extract eye gaze features and then uses fully-connected layers to recognise activities from the eye features.

\paragraph{Procedure}
We trained \textit{EHTask} using its default parameters to recognise three activities, i.e. \textit{work}, \textit{room decoration}, and \textit{meal preparation}, from the ground truth eye gaze.
At test time, we used the eye gaze generated from different methods as input to \textit{EHTask} to evaluate their effectiveness on activity recognition.

\paragraph{Results}
\autoref{tab:activity_recognition} shows the activity recognition accuracies of using the ground truth eye gaze and the eye gaze generated from different methods on the ADT dataset.
We can see that our method achieves better recognition performance than prior methods ($71.8\%$ vs. $68.7\%$) and the result is statistically significant (paired Wilcoxon signed-rank test, $p<0.01$).
We also find that the activity recognition performance of using our estimated eye gaze is comparable with that of using ground truth eye gaze ($71.8\%$ vs. $72.9\%$).
The above results demonstrate the effectiveness of using our method to estimate eye gaze for XR-related downstream tasks such as activity recognition.

\begin{table}[t] 
	\centering
	\caption{Eye-based activity recognition accuracies of different methods on ADT. Best results are in bold while the second best are underlined.} \label{tab:activity_recognition} 
        \resizebox{0.45\textwidth}{!}{
	\begin{tabular}{ccccccc}
		\toprule
		      GT & Ours &\textit{Pose2Gaze} &\textit{FixationNet} &\textit{DGaze} &\textit{Head Direction} &\textit{Chance}\\ \hline
            \textbf{72.9\%}
 &\underline{71.8\%} &68.7\% &66.6\% &66.0\% &47.1\% &33.3\%\\ 
        \bottomrule
	\end{tabular}}
\end{table}

%% file: sections/discussion.tex
\section{Discussion} \label{sec:discussion}

\paragraph{Significance of Our Method}
Our method outperforms state-of-the-art methods by an average improvement of 15.6\% for HOI setting and 6.0\% for mixed setting (\autoref{tab:results}), validating the overall superiority of our model architecture.
In addition, our method achieves superior performances than prior methods for both cross-user and cross-scene evaluations (\autoref{tab:results}), demonstrating that our method has strong generalisation capabilities for different users and different XR environments.
Furthermore, the results from the sample application of eye-based activity recognition confirm that our method can be more effective in real applications (\autoref{tab:activity_recognition}).

\paragraph{Usability of Our Method}
Our method exploits information about hand gestures and scene objects to estimate eye gaze.
\hl{Hand gesture information is readily available in many XR devices such as HTC Vive Focus 3 and Meta Quest 3 while scene object information can be directly obtained from the XR systems or easily accessed using object tracking methods~\cite{holoyolo}.}
Our method has significant potential to be integrated into such XR devices to enable numerous eye gaze-based applications including gaze-contingent rendering~\cite{patney2016towards}, gaze-based interaction~\cite{duchowski2018gaze, sidenmark2019eyehead}, or virtual content design and optimisation~\cite{alghofaili2019optimizing}.
In addition, even if dynamic hand gestures are not available, our method using static hand gestures still outperforms prior methods by a large margin, achieving an average improvement of $13.2\%$ ($9.63^\circ$ \textit{vs.} $11.10^\circ$) on HOT3D (Cross-User), $10.8\%$ ($8.74^\circ$ \textit{vs.} $9.80^\circ$) on HOT3D (Cross-Scene), and $6.0\%$ ($8.78^\circ$ \textit{vs.} $9.34^\circ$) on ADT (\autoref{tab:results} and \autoref{tab:ablation}).
These results further demonstrate the usability of our method in real applications.

\paragraph{Eye-Hand-Head Coordination}
Our key insight is that the eye, hand, and head movements are closely coordinated during HOIs and this coordination can be exploited to identify samples that are most useful for gaze estimator training -- as such, effectively denoising the training data.
Experimental results showed that using the attended hand rather than both hands and increasing the weights of coordinated eye-head training samples can significantly improve the gaze estimation performance (\autoref{tab:results}), validating the effectiveness of our insight. 
\hl{In addition, we found that the ratio of the right hand being the attended hand is 54.3\% on HOT3D and 47.1\% on ADT, demonstrating that there is no inherent bias toward the right hand as the attended hand.}

\paragraph{Limitations and Future Work}
Despite these advances, we identified several limitations that we plan to address in future work.
\hl{First, the HOT3D and ADT datasets only contain interactions with real physical objects, thus unfortunately limiting the generalisability of our evaluation.
In future work, we are looking forward to assessing our method on interactions with both real and virtual objects.}
\hl{In addtion, we exclude image/texture features from our pipeline because such features have been proven to be less effective than head movements or object positions for gaze estimation~\cite{hu2019sgaze,hu2020dgaze,hu2021fixationnet} and are computationally costly for real XR applications.
Exploring how to effectively integrate such features into our method to further boost its performance is an interesting avenue of future work.}
\hl{Finally, our method has the potential to improve the overall accuracy of eye image-based gaze estimators~\cite{zhang2017mpiigaze} by providing additional hand-head-object priors.}

%% file: sections/conclusion.tex
\section{Conclusion}

In this work, we explored the challenging task of estimating human eye gaze during hand-object interactions in extended reality.
We proposed a learning-based method that features a novel hierarchical framework, a new gaze estimator that uses CNN, GCN, and cross-modal Transformers to extract features from head movements, hand gestures, and scene objects, and a novel eye-head coordination loss.
Through extensive experiments on two public datasets, we showed that our method consistently outperforms several state-of-the-art methods by a large margin.
We also validated the effectiveness of our method for the sample application of eye-based activity recognition.
As such, our work reveals the significant information content available in eye-hand-head coordination for gaze estimation during HOIs and informs future work on this promising research direction.